\pdfoutput=1

\documentclass[11pt]{article}

\usepackage{acl}

\usepackage{times}
\usepackage{latexsym}
\usepackage[breakable]{tcolorbox}
\usepackage{amsmath}
\usepackage{booktabs}
\usepackage{adjustbox}

\usepackage[T1]{fontenc}

\usepackage[utf8]{inputenc}

\usepackage{microtype}

\usepackage{inconsolata}
\usepackage{hyperref}
\usepackage{graphicx}

\usepackage{comment}
\usepackage{xcolor, soul}
\usepackage{linguex}
\usepackage{dblfloatfix}
\title{Leveraging Human Production-Interpretation Asymmetries \\to Test LLM Cognitive Plausibility}

\author{Suet-Ying Lam$^{1*}$, Qingcheng Zeng$^{2}$\thanks{Both authors contributed equally. Correspondence to \texttt{qcz@u.northwestern.edu}}, Jingyi Wu$^{2}$, Rob Voigt$^{2}$\\\\
  $^{1}$UMass Amherst, $^{2}$Northwestern University}

\begin{document}
\maketitle
\begin{abstract}
Whether large language models (LLMs) process language similarly to humans has been the subject of much theoretical and practical debate. We examine this question through the lens of the production-interpretation distinction found in human sentence processing and evaluate the extent to which instruction-tuned LLMs replicate this distinction. Using an empirically documented asymmetry between pronoun production and interpretation in humans for implicit causality verbs as a testbed, we find that some LLMs do quantitatively and qualitatively reflect human-like asymmetries between production and interpretation. We demonstrate that whether this behavior holds depends upon both model size-with larger models more likely to reflect human-like patterns and the choice of meta-linguistic prompts used to elicit the behavior. Our codes and results are available \href{https://github.com/LingMechLab/Production-Interpretation_Asymmetries_ACL2025}{here}.
\end{abstract}

\section{Introduction and Background}
The extent to which large language models (LLMs) are ``cognitively plausible," that is, replicate human-like behaviors in language processing, has been the subject of ongoing debate \cite{doi:10.1073/pnas.2309583120, doi:10.1073/pnas.2400917121, futrell2025linguisticslearnedstopworrying, kuribayashi2025largelanguagemodelshumanlike}. Existing research on the linguistic capabilities of LLMs has predominantly focused on their performance in language interpretation, e.g., pragmatic understanding \cite{hu-etal-2023-fine}, sentence acceptability judgment \cite{warstadt2020blimp}, garden-path effect \cite{futrell-etal-2019-neural}, reference resolution \cite{lam-etal-2023-large}. In this study, we examine a previously unexplored dimension of cognitive plausibility: whether LLMs reflect human-like distinctions between \textbf{production} and \textbf{interpretation} in language processing. 


Production and interpretation were traditionally treated as two independent processes in human language: for instance, in neurolinguistics the ``classic” Lichtheim–Broca–Wernicke model assumes distinct anatomical pathways associated with production and interpretation (see \citealt{ben2008functional}). While this extreme dichotomy has been rejected recently (see \citealt{pickering2013integrated}), humans do exhibit different underlying biases in language processing between production and interpretation even in very similar tasks. Whether such distinctions carry over into LLMs is of particular interest when we consider that the fundamental unit of LLM computation is $P (token|context)$, which is inherently ambiguous between production and interpretation and is practically applied towards both types of tasks.

The present study probes into this question using reference processing as a test case. Consider the following examples:
\ex. A production task: next-mentioned bias \label{ex:free} 
\a. John infuriated Bill. ... [IC1]\label{ex:free:ic1}
\b. John praised Bill. ... [IC2]\label{ex:free:ic2}

\ex.\label{ex:pro} A interpretation task: ambiguous pronoun resolution
\a. John infuriated Bill. He ... [IC1]
\b. John praised Bill. He ...    [IC2]\label{ex:pro:ic2}

When asked to continue the story in \ref{ex:free}, speakers usually describe events that happened to one of the two mentioned characters. This is a production task, and psycholinguistic research has investigated how the preceding context affects the \textbf{next-mention bias} of the character, i.e., how likely a character will be referred to in the continued story $P(referent|context)$. In \ref{ex:free:ic1}, `John' has a higher next-mention bias than `Bill', because he is the implicit cause of the event. Verbs like `infuriate' are therefore called the subject-biased implicit causality (IC1) verbs. In contrast, `Bill' has a higher next-mention bias than `John' in \ref{ex:free:ic2}, because `Bill' implies an implicit cause for `John' to praise him. Verbs like `praise' are therefore called object-biased implicit causality (IC2) verbs (e.g., \citealt{stevenson1994thematic}). 

A similar implicit causality bias can also be found in \ref{ex:pro}. Although participants are asked to perform the same sentence continuation task as in \ref{ex:free}, note that an additional interpretation element is involved here:  since an additional pronoun is provided after the context sentence,
participants must first resolve the ambiguous pronoun ``he'' $P(referent|pronoun)$ before providing a reasonable continuation. 

Studies have shown that human participants were more likely to resolve the ambiguous pronoun to the subject `John' than the object `Bill' with an IC1 verb. That is, `John' has a higher interpretation bias than `Bill' under this condition. This interpretation bias flips when the verb is an IC2 verb: `Bill' were more likely to be the antecedent of the ambiguous pronoun `he' than `John' in \ref{ex:pro:ic2} (e.g., \citealt{crinean2006implicit}). 

Given the additional interpretation element, henceforth we refer to the sentence continuation task with a given pronoun such as \ref{ex:pro} as an \textbf{interpretation task} for measuring pronoun interpretation bias; and the one without a given pronoun such as \ref{ex:free} as the \textbf{production task} for measuring next-mentioned bias, despite the fact that the sentence continuation task is in principle a production task.

Crucially, psycholinguistic research has revealed an asymmetry between these two biases. In the interpretative case of pronoun interpretation bias, humans are robustly more likely to show a preference for the subject than in the production case of next-mention bias cross-linguistically (English: \citealt{rohde2014grammatical}; Mandarin: \citealt{zhan2020pronoun, lam2024pronoun}; German: \citealt{patterson2022bayesian}; Catalan: \citealt{mayol2018asymmetries}). That is, participants were more likely to resolve an ambiguous pronoun towards the subject than choose the subject as the next referent, despite the same context.

For humans, this extra subject bias in interpretation comes from the bias of using pronouns for subject antecedent, i.e., $P(he|subject)$. That is, when they see an ambiguous pronoun, they do not only consider the next-mention bias of which antecedent is more likely to be mentioned, but also why a pronoun is used. It is unknown whether and how LLMs can handle this difference, as they do not generate such a conditional probability based on the choice of next referent instead of the context.

We therefore probe this dimension of LLM cognitive plausibility using this task, asking (1) whether the IC verb-type effect is reflected by LLMs in both production and interpretation; and (2) whether a human-like asymmetry between the two biases exists. This is our first set of questions: do LLMs show human-like interpretation and production biases, and if so under what conditions? Do human-like effects scale with parameter count?


\paragraph{Evaluating LLM in metalinguistic prompts} 
\citet{hu-levy-2023-prompting} demonstrated that direct probability-based measures in general outperformed meta-linguistic prompting in assessing plausibility and syntactic processing tasks. However, not all language processing tasks can be effectively quantified using probability-based measures, and for some tasks meta-linguistic prompts are the only possible method to measure processing. This is exactly our case: in the ambiguous pronoun resolution task, the bias towards the subject `John' or the object 'Bill' is represented by the same term, i.e., $P(he|context)$. Metalinguistic prompting is thus necessary to elicit meaningful results. 

One might expect LLM performance to vary within different metalinguistic prompts, but it is unclear which type of metalinguistic prompts would perform better. Although \citet{hu-levy-2023-prompting} demonstrated that metalinguistic prompts that are more similar to the direct probability baseline perform better, we would like to explore whether probability measurements obtained via metalinguistic prompting also provide greater reliability. This constitutes the second aim of this paper: across different metalinguistic prompting strategies, which elicit more human-like language processing behaviors? 

Our findings are summarized below: (1) in the most cases, LLMs cannot capture the difference between the production and the interpretation task in reference processing; if any, the asymmetry is limited to certain meta-linguistic prompts and rarely reach the magnitude found in human participants; and (2) the choice of meta-linguistic prompts matters in evaluating LLMs: most LLMs perform in a more human-like manner with one specific prompt that is unrelated to probability measure, namely the Yes/No prompt (details explained in Section \ref{prompt}).

\section{Methodology}
\subsection{Stimuli}
We constructed stimuli in the frame of ``[Character A] IC-verb [Character B]" without pronouns for the next-mention production task (as in \ref{ex:free}), and with a pronoun with an ambiguous referent for the pronoun resolution task (as in \ref{ex:pro}). 
We selected 137 IC1 verbs and 134 IC2 verbs from the original study that found the production-interpretation asymmetry \cite{rohde2014grammatical}, and the English IC verb corpus \cite {ferstl2011implicit}. For each verb, we heuristically created two items by assigning a pair of male names and a pair of female names randomly selected from 13 unambiguously male and 13 unambiguously female names from \citet{rohde2008coherence}. The congruence of the gender of the characters ensures the ambiguity of the pronoun. This results in 541 items in each task.

\subsection{Models and Metalinguistic Prompting} \label{prompt}
Our experiments evaluate four representative LLMs from open-source LLMs ranging from 8B to 70B and one proprietary LLM: \texttt{LLaMA3.1Instruct-8B} \cite{grattafiori2024llama3herdmodels}, \texttt{QWen2.5Instruct-32B} \cite{qwen2025qwen25technicalreport}, \texttt{LLaMA3.3Instruct-70B} \cite{grattafiori2024llama3herdmodels}, and \texttt{GPT-4o} \cite{openai2024gpt4ocard}. We focus on instruction-tuned model \footnote{We also experiment with base LLMs. However, even for continuation prompting, base LLMs could not follow instructions reliably.} as they allow the effective use of metalinguistic prompting, and varying parameter counts also allow us to assess the impact of model scaling on human-like language processing behavior. Specifically, we constantly use greedy decoding in our generation. We employ four metalinguistic prompt strategies to assess LLM behavior:
\begin{itemize}
\setlength\itemsep{0.1em}
\item [(i)] Binary choice prompting: The model is prompted to select between subject and object.
\item [(ii)] Continuation prompting: The model is instructed to extend the sentence by continuing with either the subject or the object.
\item [(iii)] Yes/No prompting: The model is asked whether the following sentence (or the existing pronoun) will begin with the \textit{subject}, requiring a binary response.
\item [(iv)] Yes/No probability prompting: Similar to (iii), but instead of a categorical response, we extract the probability assigned to the \textit{Yes} token as a quantitative measure.

\end{itemize}
Three authors of the paper manually verified all model outputs to confirm subject/object choice and exclude ambiguous, nonsensical, plural responses, and responses with repeated pronoun in the interpretation task. Table \ref{table:exclusion} reported the distribution of the excluded responses over the 1082 responses of each model in the two tasks.

\begin{table}[h!]
\caption{The distribution of excluded responses in continuation prompting}
\label{table:exclusion}
\resizebox{\linewidth}{!}{%
\begin{tabular}{lllll}
\hline
                   & \textbf{Ambiguous} & \textbf{Non-sensical} & \textbf{Plural} &\textbf{Repeated pronoun}\\\hline
\textbf{LLaMA-8B}  & 54                 & 21                    & 17      & 0        \\
\textbf{LLaMA-70B} & 35                 & 9                     & 0       & 0        \\
\textbf{Qwen}      & 17                 & 6                     & 0    & 0           \\
\textbf{GPT-4o}      & 22                 & 1                     & 8       & 5        \\\hline
\end{tabular}}
\end{table}
The specific prompts used in our experiments are provided in Appendix~\ref{sec:prompt}, and the details of the exclusion criteria in Appendix~\ref{appendix:annotation}.

\begin{figure*}[!ht]
    \centering
    \includegraphics[width=\linewidth]{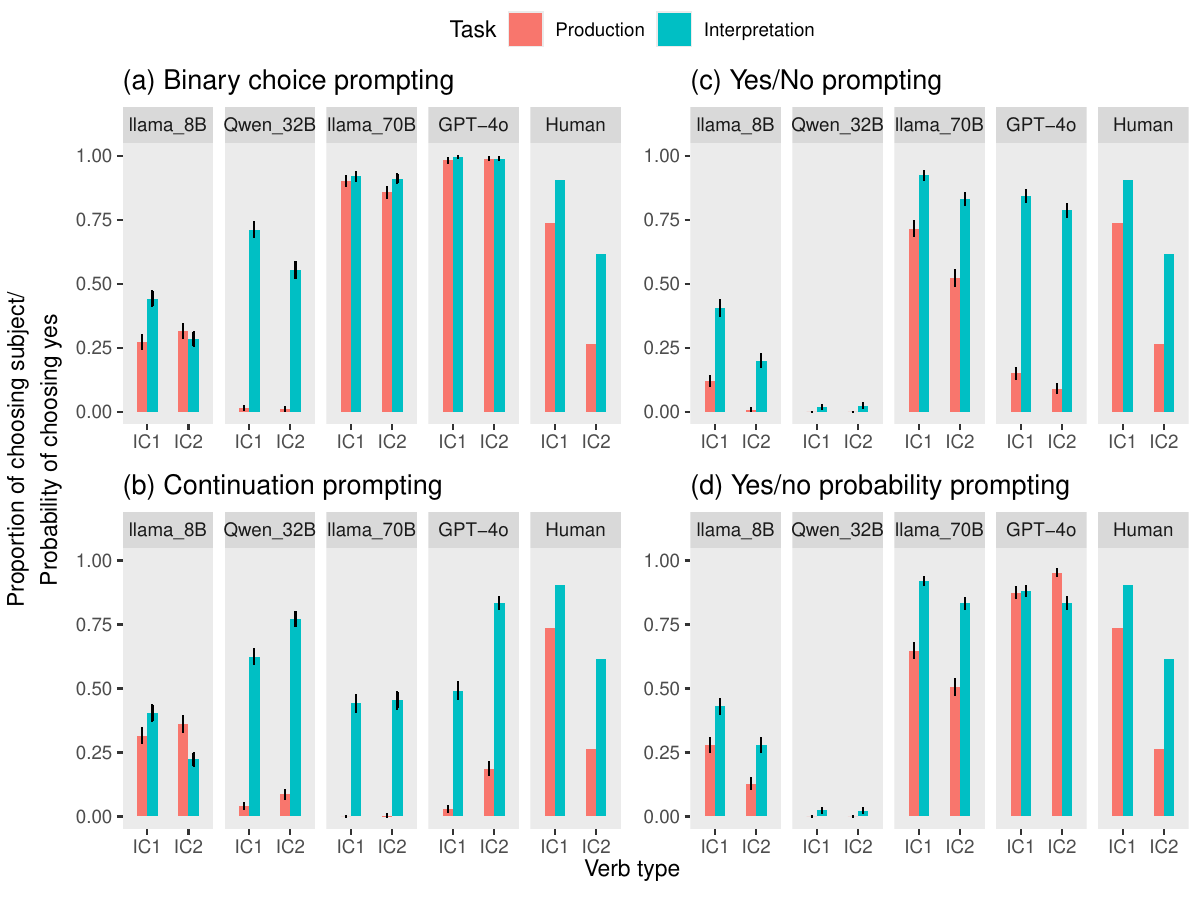}
    \caption{Model behavior as proportion of subject/yes choice as the antecedent by prompting strategy.
    Human behavior -- rightmost facet in each subfigure for reference, from \citet{rohde2014grammatical} -- tends to reflect higher subject bias for IC1 over IC2 verbs, but with asymmetry between production and interpretation tasks.}
    \vspace{-1.5em}
    \label{fig:result}
\end{figure*}

\subsection{Evaluation}
We evaluated model behavior from two perspectives. We first consider whether the IC verb type effect and the production-interpretation asymmetry exists in LLMs, which we both visualize to observe broad trends and verify with statistical tests like those run in human experiments. We then consider the magnitude of the effect found in LLMs, as psycholinguistic research has found that language models often fail to replicate the magnitude of effects found in human participants even when the directionality is similar. Below we only report effects that are verified by statistical evidence, the details of which be found in Appendix~\ref{appendix:stat}. 

\section{Results}

\paragraph{Result \#1: Implicit causality biases are sometimes replicated by LLMs, depending on prompts and models.} 
Figure \ref{fig:result} presents the task performance across different metalinguistic prompts for each model. The implicit causality (IC) verb effect — where subjects are chosen more frequently after IC1 verbs than IC2 verbs — was observed in at least one metalinguistic prompt for all models, though the strength and consistency of this effect varied by both model and prompt type.

For LLaMA models, the IC verb effect was present in both production and interpretation when using Yes/No and Yes/No probability prompts, indicating a broader sensitivity to IC biases across tasks. In contrast, the IC verb effect was limited to Yes/No prompting for GPT-4o and even more restricted for \texttt{Qwen}, in which the effect was only observed in interpretation and limited to binary choice prompt. This suggests that the model's sensitivity to IC verbs may depend on task framing. 

\paragraph{Result \#2: The production-interpretation asymmetry is limitedly captured.} Recall that human participants are more subject-biased in interpretation than in production. This difference is only elicited in LLaMA models using Yes/No and Yes/No probability prompts, and GPT-4o using Yes/No prompt and continaution prompt. \texttt{Qwen} was entirely unable to capture this asymmetry because it failed to predict the IC verb effect in production with all four metalinguistic prompts.

Another unexpected production-interpretation asymmetry is that LLMs generally are less likely to capture the IC verb bias in production than in interpretation. For instance, while the \texttt{LLaMA-8B} model was able to predict an IC verb effect in interpretation when using binary choice and continuation prompts, it predicted a reverse verb type effect in production using these prompts. The same pattern can also be seen in GPT-4o using Yes/No probability prompts. This unexpected asymmetry is even clearer in the performance of the \texttt{Qwen} model, in which the IC verb effect is only found in interpretation. This suggests that LLMs are generally unable to recognize the difference between production and interpretation like humans.

\paragraph{Result \#3: LLMs do not align with human behavior in effect magnitude.} Although there was a verb type effect and a production/interpretation task effect on LLMs' responses, the magnitude of these effects is different from humans even in the most similar case: with Yes/No prompting, \texttt{LLaMA-70B} predicted an IC verb effect difference of 19.4\% in the production task and 9.2\% in the interpretation task. Yet, these two differences were 47.2\% and 28.8\% respectively in human participants \cite{rohde2014grammatical}. In other words, even in the best scenario, LLMs underestimate the magnitude of the IC verb effect.

Surprisingly, the magnitude of the production-interpretation asymmetry predicted by LLMs is well-aligned with humans in one case: in human participants, this asymmetry was 16.8\% and 35.2\% respectively for IC1 and IC2 verbs. With Yes/No prompting, \texttt{LLaMA-70B} predicted a 20.8\% and 31\% difference respectively for the two verb types. Yet, this seems to be an isolated case. Among the six cases (two \texttt{LLaMA} models in Yes/No and Yes/No probability promptings; \texttt{GPT-4o} in Yes/No and continaution promptings) that exhibit a trend of the production-interpretation asymmetry, this is the only scenario that the magnitude of the asymmetry is similar between LLM and humans. We therefore do not tend to generalize that LLM aligns with human behavior in terms of effect magnitude. 

\paragraph{Result \#4: Scaling matters.}
Overall, \texttt{LLaMA-70B} shows better performance than \texttt{LLaMA-8B} across all four prompts. With Binary choice prompts, \texttt{LLaMA-8B} predicted a reverse IC verb effect in the production task and a reverse production-interpretation asymmetry in IC2 verbs, while both effects were captured by \texttt{LLaMA-70B}. With Yes/No and Yes/No probability prompts, although both \texttt{LLaMA-8B} and \texttt{LLaMA-70B} reflected an IC verb effect and production-interpretation asymmetry, the magnitude in human participants was better approximated by \texttt{LLaMA-70B}. 

\paragraph{Result \#5: Overall, Yes/No prompts align the best with human performance.} Among the four promptings used in our study, most models best align with human performance with the Yes/No prompting. This is the only prompting that \texttt{LLaMA} models and \texttt{GPT-4o} capture the IC verb effect and the production-interpretation asymmetry. However, for \texttt{QWen}, the best performing prompt is the binary choice prompting. This shows that models still differ in their sensitivity to metalinguistic prompts.

\section{Discussion}
Our study examines the asymmetry between interpretation and interpretation in humans within the context of LLMs, showing that, under specific prompting strategies, certain LLMs can approximate human-like asymmetry. Among the four models we tested, while \texttt{LLaMA} models and \texttt{GPT-4o} performed the best with the Yes/No prompting, \texttt{Qwen} performed the best with the binary choice prompting.

Our findings partially align with \citet{hu-levy-2023-prompting}, which suggests that the format of metalinguistic prompts influences the behavior of the model. In our study, the four distinct prompting strategies did lead to considerable variation in LLM performance, highlighting the sensitivity of model responses to prompt formulation. The different preferences for prompting strategies between model families also reinforce the need for extensive experimentation across multiple prompting approaches before evaluating a model using metalinguistic prompts.

Yet, different from their findings, we did not see prompting strategies that are more similar to direct probability measures better capture human preferences. Based on \citet{hu-levy-2023-prompting}'s observations, one might expect that metalinguistic prompts that incorporate explicit probability measures would be more reliably approximate human biases. However, our results show otherwise: although models differ in their best-performing prompts, the Yes/No prompting without probability performed best in three out of the four models in our setting. The probability-based format has never yielded superior alignment with human responses. This indicates that probability cues are not universally beneficial in prompting and that their effectiveness may depend on the specific task or model.

Another worth mentioning pattern in our results is that continuation prompting keeps performing the worst in capturing human-like performance, despite being the task originally performed by human participants. One possible explanation is that instruction-tuned LLMs, having been fine-tuned with instruction data or preference optimization objectives, may develop constrained response patterns \cite{lin2023unlockingspellbasellms}, limiting their flexibility in generating diverse continuations. Additionally, such fine-tuning can reduce conceptual diversity \cite{murthy2024fishfishseaalignment}, making LLMs less sensitive to implicit biases in language processing. This suggests that continuation prompting may not be well-suited for probing human-like asymmetries in interpretation and production.

In conclusion, our study reveals that while some instruction-tuned LLMs can approximate human-like asymmetries between production and interpretation under specific conditions, their sensitivity and alignment to human behavior are highly dependent on model scale and the choice of metalinguistic prompting strategy. These findings highlight the nuanced and prompt-contingent nature of evaluating cognitive plausibility in LLMs.

\section*{Limitations}
Our work has two primary limitations. First, our experiments are conducted on three English-centric LLMs (\texttt{LLaMA} and \texttt{GPT-4o}) and one multilingual-oriented LLM (\texttt{Qwen}). This selection may introduce biases into the models’ performance, potentially limiting the generalizability of our findings across other LLMs. Second, our study focuses solely on the asymmetry between ambiguous pronoun resolution and production in English, without exploring cross-linguistic variations. Future research could address these limitations by incorporating a more diverse set of LLMs and broadening the scope of languages analyzed. 

\bibliography{custom}

\appendix

\section{Prompt Template}
\label{sec:prompt}
The following shows the metalinguistic prompt design for LLMs.
\begin{tcolorbox}[colback=gray!10!white, colframe=gray!50!gray, halign=left, top=0mm, bottom=0mm, left=1mm, right=1mm, boxrule=0.5pt, breakable]
\fontsize{9pt}{9pt}\selectfont 
\linespread{1}\selectfont
System Prompt:  You are a helpful assistant.\\
\vspace{10pt}
production-binary-choice-template: \\
In the following sentence, who is more likely to be the subject of the next sentence? \{\} or \{\}? Please ONLY return the name without any explanation or extra words.\\
Sentence: \{\}\\
Answer:\\
\vspace{10pt}
production-yes-no-template: \\
In the following sentence, judge whether the pronoun of the next sentence will refer to \{\}. Please ONLY answer with 'Yes' or 'No'.\\
Sentence: \{\}\\
\vspace{10pt}
production-continuation-template: \\
Please reasonably continue the sentence with one of the mentioned characters. You should start a new sentence rather than a clause. Please ONLY return the continuation.\\
Sentence: \{\}\\
\vspace{10pt}
interpretation-binary-choice-template:\\
In the following sentence, who is more likely to be the referent of the pronoun? \{\} or \{\}? Please ONLY return the name without any explanation or extra words.\\
Sentence: \{\}\\
Answer:\\
\vspace{10pt}
interpretation-yes-no-template:\\
In the following sentence, judge whether the pronoun refers to \{\}. Please ONLY answer with 'Yes' or 'No'.\\
Sentence: \{\}\\
\vspace{10pt}
interpretation-continuation-template:\\
Please reasonably continue the sentence following the pronoun. Please ONLY return the continuation.
Sentence: \{\}
\end{tcolorbox}

\section{Statistical analyses and results} 
\label{appendix:stat}
We focus on how the bias of IC verbs (IC1 vs. IC2) and the task (production vs. interpretation) affect the outcomes of LLMs in each type of meta-linguistic prompts.

\subsection{Data annotation} \label{appendix:annotation}
For binary prompts, we annotated based on the name answered by the LLM. For yes-no prompts, we annotated "yes" as ``Subject" .

For continuation prompts, we manually annotated the choice of referent (Subject vs. Object) in the production task and the choice of antecedent (Subject vs. Object) in the ambiguous pronoun resolution task based on the meaning of the generated continuation. For instance, for a sentence like ``Nick offended Steve. He decided to apologize and clear the air before things escalated further", we annotated the outcome as ``Subject". Ambiguous references (e.g., ``Zack divorced Paul. He later moved to a new city to start his life over."), non-sensical continuations (e.g., ``Janet wanted Kate. She to join her at the party that night, but Kate had already made other plans."),  continuations with a plural antecedent (e.g., ``Claire played with Jane. They were building a sandcastle on the beach"), and continuations starting with a repeated pronoun in an interpretation task (e.g., ``Claire played with Jane. She... she") were excluded. The distribution of the excluded item has been provided in Table \ref{table:exclusion} in the main text, repeated below:

\begin{table}[h!]
\caption{The distribution of excluded responses in continuation prompting}
\label{table:exclusion}
\resizebox{\linewidth}{!}{%
\begin{tabular}{lllll}
\hline
                   & \textbf{Ambiguous} & \textbf{Non-sensical} & \textbf{Plural} &\textbf{Repeated pronoun}\\\hline
\textbf{LLaMA-8B}  & 54                 & 21                    & 17      & 0        \\
\textbf{LLaMA-70B} & 35                 & 9                     & 0       & 0        \\
\textbf{Qwen}      & 17                 & 6                     & 0    & 0           \\
\textbf{GPT-4o}      & 22                 & 1                     & 8       & 5        \\\hline
\end{tabular}}
\end{table}

As can be seen, \texttt{LLaMA-8B} made more ambiguous and non-sensical continuations than all other models. Besides, it also provided continuations with ``they", which violates the requirement posed in the prompts. This might suggest that scaling affects LLMs' ability to follow the instructions. More models should be evaluated to test this hypothesis.

\subsection{Analyses}
\paragraph{LLaMA models}
For the results of LLaMA models, we ran a mixed-effects Bayesian bernoulli regression model using the R package \texttt{brms} for the binary outcome resulted from continuation, binary, and yes-no prompts (Subject = 1; Object = 0); and a mixed-effects Bayesian linear regression model for the continuous probability outcome from yes-no probability prompt. Note that the dataset used in statistical model for continuation outcome is slightly different from that used in other models, as some of the responses are excluded, as introduced in Appendix \ref{appendix:annotation}.

Each model was fitted using 4 chains, each with 5000 iterations. The first 1000 were warm-up to calibrate the sampler. This results in 12000 posterior samples. They were all built with fixed predictors of IC verbs (sum-coded: IC1 = 0.5; IC2 = -0.5), task type (sum-coded: interpretation = 0.5; production = -0.5), and their interaction. A maximal random structure justified by design is implemented \cite{barr2013random}. For logistic regression models, we used weakly informative priors, i.e., a Cauchy distribution with a center of 0 and a scale of 2.5 for fixed effects following \citet{gelman2008weakly}, and the default setting of the package for the other parameters. For linear regression models, we used a gaussian distribution with a mean of 0 and a standard deviation of 1 as the weakly informative prior for fixed predictors. When there is an interaction effect, we further ran nested models for pairwise comparison.

The Bayesian statistics framework does not use the p-value. We consider the 95\% credible interval (Crl) as the evidence for an effect: if the 95\% Crl does not include a zero, i.e., it is all positive or negative, we consider there is evidence for an effect. Below we report the estimate and the 95\% Crl for each effect.

\paragraph{QWen} For \texttt{Qwen}, we only ran analysis for continuation prompts, and limited to the effect of the IC verb only. This is because responses of \texttt{Qwen} in the production task is so extreme that no statistical model can be successfully fitted. The settings of the mixed-effects Bayesian bernoulli regression model are the same as those used for LLaMA models.

\paragraph{GPT-4o} Like LLaMA models, we ran mixed-effects Bayesian bernoulli regression model for binary outcomes from continuation and yes-no prompts. We did not run analysis for binary prompts because responses of \texttt{GPT-4o} were extremely subject-biased in all conditions. As a result, no statistical model can be fitted. We also ran linear regression model for probability outcomes from yes-no probability prompts. The settings of the statistical models are the same as in analyses for LLaMA models.

\subsection{Results} We bold the predictor in which the effect is supported the statistical evidence, i.e., the 95\% Crl does not contain a zero.
\subsubsection{LLaMA-3.1-8B Model}
\paragraph{Binary prompting} Table 3 shows that there was an interaction effect between the IC verb type and the task type. 

\begin{table}[h]
    \centering
    \resizebox{\columnwidth}{!}{
    \begin{tabular}{lccc}
        \toprule
        \multicolumn{4}{c}{\textbf{Formula:} $\text{binary} \sim \text{verb} * \text{task} + (1 + \text{task} | \text{itemID})$} \\
        \midrule
        & Estimate & Est. Error & 95\% CrI \\
        \midrule
        Intercept & -1.80 & 0.74 & \textbf{[-3.78, -0.99]} \\
        verb & 0.64 & 0.48 & [-0.12, 1.80] \\
        task & 0.67 & 0.75 & [-0.65, 2.47] \\
        \textbf{verb:task} & 2.06 & 0.99 & \textbf{[0.78, 4.61]} \\
        \bottomrule
    \end{tabular}}
    \caption{Summary of the Bayesian logistic regression model of LLaMA-3.1-8B model, binary choice prompting.}
\end{table}

Nested analyses further reveal that the production-interpretation asymmetry is only found in IC1 verbs (Table 4), and the verb type effect is only found in interpretation (Table 5).

\begin{table}[h]
    \centering
    \resizebox{\columnwidth}{!}{
    \begin{tabular}{lccc}
        \toprule
        \multicolumn{4}{c}{\textbf{Formula:} $\text{binary} \sim \text{verb} / \text{task} + (1 + \text{task} | \text{itemID})$} \\
        \midrule
        & Estimate & Est. Error & 95\% CrI \\
        \midrule
        \textbf{Intercept} & -1.84 & 0.76 & \textbf{[-3.81, -0.99]} \\
        verb & 0.63 & 0.47 & [-0.14, 1.76] \\
        \textbf{verbIC1:task} & 1.76 & 0.96 & \textbf{[0.55, 4.24]} \\
        verbIC2:task & -0.40 & 0.82 & [-2.30, 1.14] \\
        \bottomrule
    \end{tabular}}
    \caption{Pairwise comparison of the task type effect within IC1 and within IC2 conditions using binary choice prompting in LLaMA-3.1-8B model.}
\end{table}

\begin{table}[h]
    \centering
    \resizebox{\columnwidth}{!}{
    \begin{tabular}{lccc}
        \toprule
        \multicolumn{4}{c}{\textbf{Formula:} $\text{binary} \sim \text{task} / \text{verb} + (1 + \text{prompt} | \text{itemID})$} \\
        \midrule
        & Estimate & Est. Error & 95\% CrI \\
        \midrule
        \textbf{Intercept} & -1.80 & 0.79 & \textbf{[-3.85, -0.99]} \\
        task & 0.65 & 0.73 & [-0.68, 2.37] \\
        taskProduction:verb & -0.43 & 0.50 & [-1.57, 0.42] \\
        \textbf{taskInterpretation:verb} & 1.69 & 0.89 & \textbf{[0.61, 4.01]} \\
        \bottomrule
    \end{tabular}}
    \caption{Pairwise comparison of the IC verb type effect
within the production and within the interpretation task using binary
choice prompting in LLaMA-3.1-8B model.}
\end{table}

\newpage
\paragraph{Yes-no prompting} As shown in Table 6, there was an interaction effect between the verb type and task type for the responses of the model. 

\begin{table}[h]
    \centering
    \resizebox{\columnwidth}{!}{
    \begin{tabular}{lccc}
        \toprule
        \multicolumn{4}{c}{\textbf{Formula:} $\text{yes\_no} \sim \text{verb} * \text{task} + (1 | \text{itemID})$} \\
        \midrule
        & Estimate & Est. Error & 95\% CrI \\
        \midrule
        \textbf{Intercept} & -4.06 & 0.61 & \textbf{[-5.44, -3.04]} \\
        \textbf{verb} & 3.12 & 0.67 & \textbf{[1.97, 4.61]} \\
        \textbf{task} & 4.31 & 0.70 & \textbf{[3.12, 5.89]} \\
        \textbf{verb:task} & -1.99 & 0.93 & \textbf{[-4.01, -0.39]} \\
        \bottomrule
    \end{tabular}}
    \caption{Summary of the Bayesian logistic regression model of LLaMA-3.1-8B model, yes-no prompting.}
\end{table}

Nested analysis in Table 7 further reveals that the model did give more "yes" (or referring to subject) with IC1 verbs than IC2 verbs (as indicated by the positive intercept of the verb predictor). Also, the production-interpretation asymmetry is found within both IC1 and IC2 verbs, such that the model chose more subjects in interpretation than production.

\begin{table}[h]
    \centering
    \resizebox{\columnwidth}{!}{
    \begin{tabular}{lccc}
        \toprule
        \multicolumn{4}{c}{\textbf{Formula:} $\text{yes\_no} \sim \text{verb} / \text{task} + (1 | \text{itemID})$} \\
        \midrule
        & Estimate & Est. Error & 95\% CrI \\
        \midrule
        \textbf{Intercept} & -4.06 & 0.62 & \textbf{[-5.50, -3.04]} \\
        \textbf{verb} & 3.18 & 0.70 & \textbf{[2.00, 4.76]} \\
       \textbf{verbIC1:task} & 3.22 & 0.55 & \textbf{[2.28, 4.43]} \\
        \textbf{verbIC2:task} & 5.45 & 1.14 & \textbf{[3.58, 8.04]} \\
        \bottomrule
    \end{tabular}}
    \caption{Pairwise comparison of the task effect
within the IC1 and within the IC2 verbs using yes-no choice prompting in LLaMA-3.1-8B model.}
\end{table}

\paragraph{Continuation prompting} As shown in Table 8, there was an interaction effect between the verb type and the task type for the responses of the model. 

\begin{table}[h]
    \centering
    \resizebox{\columnwidth}{!}{
    \begin{tabular}{lccc}
        \toprule
        \multicolumn{4}{c}{\textbf{Formula:} $\text{cont} \sim \text{verb} * \text{task} + (1 | \text{itemID})$} \\
        \midrule
        & Estimate & Est. Error & 95\% CrI \\
        \midrule
        \textbf{Intercept} & -0.81 & 0.09 & \textbf{[-0.99, -0.65]} \\
        \textbf{verb} & 0.36 & 0.16 & \textbf{[0.05, 0.67]} \\
        task & -0.15 & 0.14 & [-0.44, 0.13] \\
        \textbf{verb:task} & 1.14 & 0.30 & \textbf{[0.57, 1.73]} \\
        \bottomrule
    \end{tabular}}
    \caption{Summary of the Bayesian logistic regression model of LLaMA-3.1-8B model, continuation prompting.}
\end{table}

Nested analyses in Table 9 show that the verb type effect is only found in interpretation. 

\begin{table}[h!]
    \centering
    \resizebox{\columnwidth}{!}{
    \begin{tabular}{lccc}
        \toprule
        \multicolumn{4}{c}{\textbf{Formula:} $\text{cont} \sim \text{task} / \text{verb} + (1 | \text{itemID})$} \\
        \midrule
        & Estimate & Est. Error & 95\% CrI \\
        \midrule
        \textbf{Intercept} & -0.81 & 0.09 & \textbf{[-0.99, -0.64]} \\
        task & -0.15 & 0.14 & [-0.44, 0.13] \\
        taskProduction:verb & -0.22 & 0.21 & [-0.64, 0.18] \\
        \textbf{taskInterpretation:verb} & 0.93 & 0.22 & \textbf{[0.51, 1.37]} \\
        \bottomrule
    \end{tabular}}
    \caption{Pairwise comparison of the IC verb type effect
within the production and within the interpretation task using continuation prompting in LLaMA-3.1-8B model.}
\end{table}

The pairwise comparison in Table 10 shows that the production-interpretation asymmetry differs in direction between the two verb types: while interpretation is more subject-biased than production in IC1 verbs, production is more subject-biased than interpretation in IC2 verbs.

\begin{table}[h!]
    \centering
    \resizebox{\columnwidth}{!}{
    \begin{tabular}{lccc}
        \toprule
        \multicolumn{4}{c}{\textbf{Formula:} $\text{cont} \sim \text{verb} / \text{task} + (1 | \text{itemID})$} \\
        \midrule
        & Estimate & Est. Error & 95\% CrI \\
        \midrule
        \textbf{Intercept} & -0.81 & 0.09 & \textbf{[-0.99, -0.65]} \\
        \textbf{verb} & 0.36 & 0.15 & \textbf{[0.06, 0.66]} \\
        \textbf{verbIC1:task} & 0.42 & 0.20 & \textbf{[0.05, 0.81]} \\
        \textbf{verbIC2:task}& -0.73 & 0.21 & \textbf{[-1.14, -0.32]} \\
        \bottomrule
    \end{tabular}}
    \caption{Pairwise comparison of the task type effect
within IC1 and within IC2 verbs using continuation prompting in LLaMA-3.1-8B model.}
\end{table}

\paragraph{Yes/No probability prompting} As shown in Tabel 11, the model did generate a higher probability of `Yes' (= subject) following IC1 verbs than IC2 verbs, and in interpretation task than in production task. 

\begin{table}[h!]
    \centering
    \resizebox{\columnwidth}{!}{
    \begin{tabular}{lccc}
        \toprule
        \multicolumn{4}{c}{\textbf{Formula:} $\text{subject\_yes\_probability} \sim \text{verb} * \text{ task} + (1 | \text{itemID})$} \\
        \midrule
        & Estimate & Est. Error & 95\% CrI \\
        \midrule
        \textbf{Intercept} & 0.28 & 0.01 & \textbf{[0.27, 0.29]} \\
        \textbf{verb} & 0.15 & 0.01 & \textbf{[0.12, 0.18]} \\
        \textbf{task} & 0.15 & 0.01 & \textbf{[0.13, 0.17]} \\
        verb: task & -0.00 & 0.02 & [-0.03, 0.03] \\
        \bottomrule
    \end{tabular}}
    \caption{Summary of the Bayesian linear regression model of LLaMA-3.1-8B model, Yes/No probability prompting.}
\end{table}

\subsubsection{LLaMA-3.3-70B}
\paragraph{Binary prompting} Table 12 clearly shows that the model only reveals the production-interpretation asymmetry, such that it chose more subject in interpretation than in production. There was no clear evidence for the IC verb type effect.

\begin{table}[h]
    \centering
    \resizebox{\columnwidth}{!}{
    \begin{tabular}{lccc}
        \toprule
        \multicolumn{4}{c}{\textbf{Formula:} $\text{binary} \sim \text{verb} * \text{task} + (1 | \text{itemID})$} \\
        \midrule
        & Estimate & Est. Error & 95\% CrI \\
        \midrule
        \textbf{Intercept} & 14.61 & 3.61 & \textbf{[9.24, 23.34]} \\
        verb & 1.36 & 1.42 & [-1.19, 4.45] \\
        \textbf{task} & 2.21 & 0.73 & \textbf{[0.98, 3.84]} \\
        verb:task & -1.24 & 1.07 & [-3.44, 0.77] \\
        \bottomrule
    \end{tabular}}
    \caption{Summary of the Bayesian logistic regression model of LLaMA-3.3-70B model, binary prompting.}
\end{table}

\paragraph{Yes-no prompting} Table 13 shows that the model is able to capture the IC verb type effect and the production-interpretation asymmetry, such that it responded `Yes' (=subject) more for IC1 verbs than IC2 verbs and the interpretation task than the production task.

\begin{table}[h]
    \centering
    \resizebox{\columnwidth}{!}{
    \begin{tabular}{lccc}
        \toprule
        \multicolumn{4}{c}{\textbf{Formula:} $\text{yes\_no} \sim \text{verb} * \text{task} + (1 | \text{itemID})$} \\
        \midrule
        & Estimate & Est. Error & 95\% CrI \\
        \midrule
        \textbf{Intercept} & 2.24 & 0.25 & \textbf{[1.79, 2.78]} \\
        \textbf{verb} & 1.46 & 0.34 & \textbf{[0.83, 2.15]} \\
        \textbf{task} & 2.63 & 0.31 & \textbf{[2.05, 3.29]} \\
        verb:task & -0.10 & 0.44 & [-0.95, 0.76] \\
        \bottomrule
    \end{tabular}}
    \caption{Summary of the Bayesian logistic regression model of LLaMA-3.3-70B model, Yes/No prompting.}
\end{table}

\paragraph{Continuation prompting} Like in binary choice prompting, Table 14 shows that the model only reveals the production-interpretation asymmetry, such that it chose more subject in interpretation than in production. There was no clear evidence for the IC verb type effect.
\begin{table}[h]
    \centering
    \resizebox{\columnwidth}{!}{
    \begin{tabular}{lccc}
        \toprule
        \multicolumn{4}{c}{\textbf{Formula:} $\text{cont} \sim \text{verb} * \text{task} + (1 | \text{itemID})$} \\
        \midrule
        & Estimate & Est. Error & 95\% CrI \\
        \midrule
        \textbf{Intercept} & -3.66 & 0.66 & \textbf{[-5.23, -2.68]} \\
        verb & -0.61 & 0.88 & [-2.67, 0.86] \\
        \textbf{task} & 6.83 & 1.27 & \textbf{[4.89, 9.82]} \\
        verb:task & 1.10 & 1.74 & [-1.78, 5.19] \\
        \bottomrule
    \end{tabular}}
    \caption{Summary of the Bayesian logistic regression model of LLaMA-3.3-70B model, continuation prompting.}
\end{table}

\paragraph{Yes/No probability prompting} Table 15 shows an interaction effect between verb and task type.
\newpage

\begin{table}[h]
    \centering
    \resizebox{\columnwidth}{!}{
    \begin{tabular}{lccc}
        \toprule
        \multicolumn{4}{c}{\textbf{Formula:} $\text{subject\_yes\_probability} \sim \text{verb} * \text{task} + (1 | \text{itemID})$} \\
        \midrule
        & Estimate & Est. Error & 95\% CrI \\
        \midrule
        \textbf{Intercept} & 0.73 & 0.01 & \textbf{[0.70, 0.75]} \\
        \textbf{verb} & 0.12 & 0.02 & \textbf{[0.07, 0.16]} \\
        \textbf{task} & 0.30 & 0.01 & \textbf{[0.27, 0.33]} \\
        \textbf{verb:task} & -0.05 & 0.03 & \textbf{[-0.11, -0.00]} \\
        \bottomrule
    \end{tabular}}
    \caption{Summary of the Bayesian linear regression model of LLaMA-3.3-70B model, Yes/No probability prompting.}
\end{table}

Pairwise comparisons in Table 16 and 17 further show that the verb type effect can be found in both production and interpretation task, and the asymmetry can be found in both IC1 and IC2 conditions.

\begin{table}[h]
    \centering
    \resizebox{\columnwidth}{!}{
    \begin{tabular}{lccc}
        \toprule
        \multicolumn{4}{c}{\textbf{Formula:} $\text{subject\_yes\_probability} \sim \text{verb} / \text{task} + (1 | \text{itemID})$} \\
        \midrule
        & Estimate & Est. Error & 95\% CrI \\
        \midrule
        \textbf{Intercept} & 0.73 & 0.01 & \textbf{[0.71, 0.75]} \\
        \textbf{verb} & 0.12 & 0.02 & \textbf{[0.07, 0.16]} \\
        \textbf{verbIC1:task} & 0.27 & 0.02 & \textbf{[0.24, 0.31]} \\
        \textbf{verbIC2:task} & 0.33 & 0.02 & \textbf{[0.29, 0.37]} \\
        \bottomrule
    \end{tabular}}
    \caption{Pairwise comparison of the verb type effect
within the production and the interpretation task using Yes/No probability prompting in LLaMA-3.1-70B model.}
\end{table}

\begin{table}[h]
    \centering
    \resizebox{\columnwidth}{!}{
    \begin{tabular}{lccc}
        \toprule
        \multicolumn{4}{c}{\textbf{Formula:} $\text{subject\_yes\_probability} \sim \text{task} / \text{verb} + (1 | \text{itemID})$} \\
        \midrule
        & Estimate & Est. Error & 95\% CrI \\
        \midrule
        \textbf{Intercept} & 0.73 & 0.01 & \textbf{[0.70, 0.75]} \\
        \textbf{task} & 0.30 & 0.01 & \textbf{[0.27, 0.33]} \\
        \textbf{taskProduction:verb} & 0.14 & 0.03 & \textbf{[0.09, 0.19]} \\
        \textbf{taskInterpretation:verb} & 0.09 & 0.03 & \textbf{[0.04, 0.14]} \\
        \bottomrule
    \end{tabular}}
    \caption{Pairwise comparison of the task type effect
within IC1 and within IC2 verbs using Yes/No probability prompting in LLaMA-3.1-70B model.}
\end{table}

\subsubsection{QWen} Note that we only analyzed the responses of the continuation prompting in the interpretation task for \texttt{Qwen}, because models cannot be converged in other case. As can be seen below in Table 18, there is an opposite IC verb type effect such that the model referred to more subjects following IC2 verbs than IC1 verbs.
\newpage

\begin{table}[h]
    \centering
    \resizebox{\columnwidth}{!}{
    \begin{tabular}{lccc}
        \toprule
        \multicolumn{4}{c}{\textbf{Formula:} $\text{cont} \sim \text{verb} + (1 | \text{itemID})$} \\
        \midrule
        & Estimate & Est. Error & 95\% CrI \\
        \midrule
        \textbf{Intercept} & -0.60 & 0.07 & \textbf{[-0.73, -0.47]} \\
        \textbf{verb} & -0.37 & 0.13 & \textbf{[-0.63, -0.11]} \\
        \bottomrule
    \end{tabular}}
    \caption{Summary of the Bayesian logistic regression model of \texttt{Qwen} model, continuation prompting.}
\end{table}

\subsubsection{GPT-4o}
Recall that due to the extreme pattern of \texttt{GPT-4o} with binary promptings, no statistical model could be fitted. Below we only report results of statistical analyses for outputs from Yes/No, Yes/No probability, and continuation promptings only.

Table 19 shows the results of outcomes from Yes/No prompting of \texttt{GPT-4o}. As can be seen, there was a main effect of both verb type and task type, such that IC1 verbs and interpretation task elicited more choices towards subject antecedent. This aligns with the human performance.

\begin{table}[h]
    \centering
    \resizebox{\columnwidth}{!}{
    \begin{tabular}{lccc}
        \toprule
        \multicolumn{4}{c}{\textbf{Formula:} $\text{yes\_no} \sim \text{verb} * \text{task} + (1 | \text{itemID})$} \\
        \midrule
        & Estimate & Est. Error & 95\% CI \\
        \midrule
        \textbf{Intercept} & -0.39 & 0.15 & \textbf{[-0.69, -0.12]} \\
        \textbf{verb} & 0.68 & 0.29 & \textbf{[0.15, 1.28]} \\
        \textbf{task} & 5.51 & 0.83 & \textbf{[4.17, 7.42]} \\
        verb:task & -0.23 & 0.46 & [-1.15, 0.67] \\
        \bottomrule
    \end{tabular}}
    \caption{Summary of the Bayesian logistic regression model of \texttt{GPT-4o}, Yes/No prompting.}
\end{table}

Table 20 shows the results of outcomes from Yes/No probability prompting of \texttt{GPT-4o}. Given statistical evidence for the interaction effect between verb and task type, we further ran a nested analysis for the effect of the task type within each level of verb, as shown in Table 11.

\begin{table}[h]
    \centering
    \resizebox{\columnwidth}{!}{
    \begin{tabular}{lccc}
        \toprule
        \multicolumn{4}{c}{\textbf{Formula:} $\text{subject\_yes\_probability} \sim \text{verb} * \text{task} + (1 | \text{itemID})$} \\
        \midrule
        & Estimate & Est. Error & 95\% CI \\
        \midrule
        \textbf{Intercept} & 0.89 & 0.01 & \textbf{[0.87, 0.90]} \\
        verb & -0.01 & 0.01 & [-0.04, 0.01] \\
        \textbf{task} & -0.06 & 0.01 & \textbf{[-0.08, -0.03]} \\
        \textbf{verb:task} & 0.13 & 0.02 & \textbf{[0.08, 0.17]} \\
        \bottomrule
    \end{tabular}}
    \caption{Summary of the Bayesian linear regression model of \texttt{GPT-4o}, Yes/No probability prompting.}
\end{table}

As can be seen, there was only a task effect limited to IC2 verbs and in an opposite way to human performance: the interpretation task was even less subject-biased than the production task, as shown in the negative sign of the estimate of verbIC2:task.

\begin{table}[h]
    \centering
    \resizebox{\columnwidth}{!}{
    \begin{tabular}{lccc}
        \toprule
        \multicolumn{4}{c}{\textbf{Formula:} $\text{subject\_yes\_probability} \sim \text{verb} / \text{prompt} + (1 | \text{itemID})$} \\
        \midrule
        & Estimate & Est. Error & 95\% CI \\
        \midrule
        Intercept & 0.89 & 0.01 & \textbf{[0.87, 0.90]} \\
        verb & -0.01 & 0.01 & [-0.04, 0.01] \\
        verbIC1:task & 0.01 & 0.02 & [-0.03, 0.04] \\
        \textbf{verbIC2:task} & -0.12 & 0.02 & \textbf{[-0.15, -0.08]} \\
        \bottomrule
    \end{tabular}}
    \caption{Pairwise comparison of the task type effect
within IC1 and within IC2 verbs using of \texttt{GPT-4o}, Yes/No probability prompting.}
\end{table}

Lastly, Table 22 shows statistical results from continuation promptings of \texttt{GPT-4o}. The main effect of the task type shows that the model captures the production-interpretation asymmetry that the interpretation task was more subject-biased than the production task. However, the negative sign of the verb type effect shows a reversed IC verb effect, such that IC2 verbs were more subject-biased than IC1 verbs. 

\begin{table}[h]
    \resizebox{\columnwidth}{!}{
    \begin{tabular}{lccc}
        \toprule
        \multicolumn{4}{c}{\textbf{Formula:} $\text{cont} \sim \text{verb} * \text{task} + (1 | \text{itemID})$} \\
        \midrule
        & Estimate & Est. Error & 95\% CI \\
        \midrule
        Intercept & -1.43 & 0.30 & \textbf{[-2.12, -0.95]} \\
        \textbf{verb} & -3.11 & 0.63 & \textbf{[-4.57, -2.10]} \\
        \textbf{task} & 5.85 & 1.04 & \textbf{[4.26, 8.28]} \\
        verb:task & 0.10 & 0.60 & [-1.05, 1.33] \\
        \bottomrule
    \end{tabular}}
    \caption{Summary of the Bayesian linear regression model of \texttt{GPT-4o}, continuation prompting.}
\end{table}

\end{document}